\documentclass[10pt,twocolumn,letterpaper]{article}

\usepackage{wacv}
\usepackage{times}
\usepackage{epsfig}
\usepackage{graphicx}
\usepackage{amsmath}
\usepackage{amssymb}
\usepackage[accsupp]{axessibility} 

\usepackage{authblk}
\usepackage{pifont}
\usepackage{mathtools}
\usepackage{booktabs}
\usepackage{multirow}
\usepackage{adjustbox}
\usepackage{rotating}
\usepackage{paralist}
\usepackage{comment}
\usepackage{bbm}
\usepackage{colortbl}
\usepackage{pbox}

\usepackage[dvipsnames]{xcolor}
\newcommand\our{\text{PixDA}}
\newcommand\ourloss{\text{PixAdv}}

\newcommand\boldred[1]{\textcolor{red}{\textbf{#1}}}
\newcommand{\myparagraph}[1]{\vspace{4pt}\noindent\textbf{#1}}
\newcommand{\cmark}{\ding{51}}

\def\wacvPaperID{568} % Enter the WACV Paper ID here

\wacvfinalcopy % *** Uncomment this line for the final submission

\ifwacvfinal
\fi

\ifwacvfinal
\usepackage[breaklinks=true, bookmarks=false, colorlinks]{hyperref}
\else
\usepackage[pagebackref=true,breaklinks=true,colorlinks,bookmarks=false]{hyperref}
\fi

\usepackage{my_ref}

\begin{document}

%%%%%%%%% TITLE
\title{Pixel-by-Pixel Cross-Domain Alignment for Few-Shot Semantic Segmentation\vspace{-15pt}}

\author[1]{Antonio Tavera}
\author[1]{Fabio Cermelli}
\author[2]{Carlo Masone}
\author[1]{Barbara Caputo\vspace{-0.5cm}}
\affil[1]{\small Polytechnic University of Turin, Turin, Italy
}
\affil[2]{\small CINI - Consorzio Interuniversitario Nazionale per l'Informatica, Rome, Italy}
\affil[1]{\{\tt\small antonio.tavera, fabio.cermelli, barbara.caputo\}@polito.it}

\maketitle
\ifwacvfinal
\thispagestyle{empty}
\fi

%%%%%%%%% ABSTRACT
\begin{abstract}
In this paper we consider the task of semantic segmentation in autonomous driving applications. Specifically, we consider the cross-domain few-shot setting where training can use only few real-world annotated images and many annotated synthetic images. In this context, aligning the domains is made more challenging by the pixel-wise class imbalance that is intrinsic in the segmentation and that leads to ignoring the underrepresented classes and overfitting the well represented ones.
We address this problem with a novel framework called Pixel-By-Pixel Cross-Domain Alignment ({\our}). 
We propose a novel pixel-by-pixel domain adversarial loss following three criteria: (i) align the source and the target domain for each pixel, (ii) avoid negative transfer on the correctly represented pixels, and (iii) regularize the training of infrequent classes to avoid overfitting.
The pixel-wise adversarial training is assisted by a novel sample selection procedure, that handles the imbalance between source and target data, and a knowledge distillation strategy, that avoids overfitting towards the few target images.
We demonstrate on standard synthetic-to-real benchmarks that {\our} outperforms previous state-of-the-art methods in (1-5)-shot settings.\footnote{Code at:  \url{https://github.com/taveraantonio/PixDA}.}
\end{abstract}

%%%%%%%%% BODY TEXT
\begin{figure}[ht]
\begin{center}
\includegraphics[width=1.0\columnwidth]{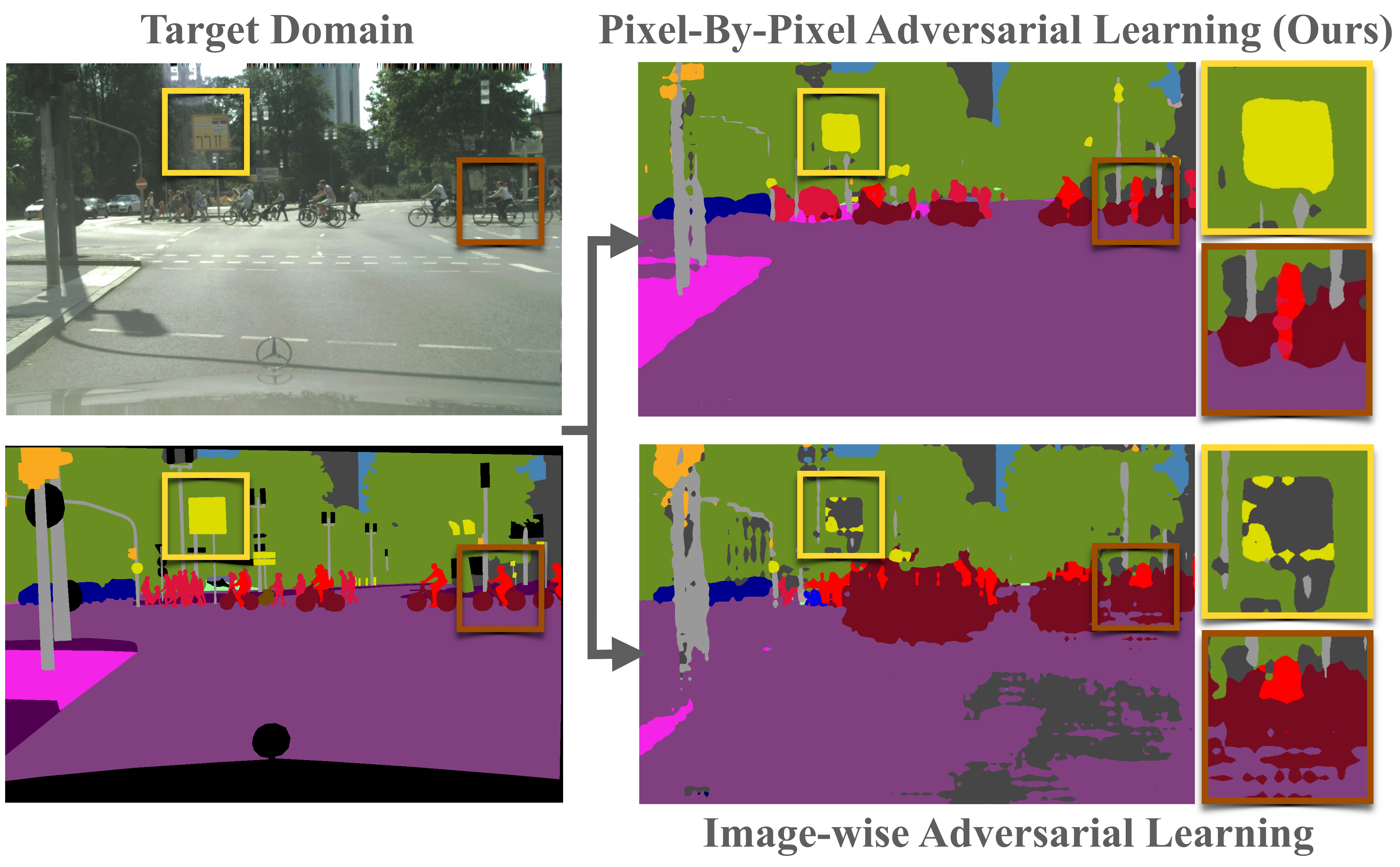}
\end{center}
\caption{
Comparison between the common image-wise adversarial training (bottom) and {\our} that analyzes each pixel individually (top). By prioritizing the pixel alignment according to the imbalance of the classes and the network classification confidence, {\our} achieves better accuracy, particularly for underrepresented semantic classes, \eg traffic sign, rider and bicycle.
}
\label{fig:teaser}
\vspace{-10pt}
\end{figure}

\section{Introduction}
Semantic segmentation is a foundational technology in autonomous driving applications, because it provides the vehicle with information about its surroundings that is critical to safely and reliably navigate the environment.
Great strides have been made to improve this technology in the autonomous driving scenario, using supervised deep learning methods trained using great quantities of data with pixel-wise annotations.

However, data collection and annotation is time consuming, expensive and hard-to-scale.
A successful strategy to mitigate this issue is to rely on simulators to generate massive amounts of synthetic data \cite{Richter2016GTA, Ros2016Synthia, tavera2020idda}.
This solution has the benefit that synthetic data is easy and cheap to collect, and the semantic annotations generated automatically by the graphical engines are perfect. The downside is that there is a significant shift between the synthetic domain of the training data and the real domain of the application.
There are unsupervised and semi-supervised solutions that address this domain shift \cite{survey_uda1, survey_uda2}, however they still need large amounts of images from the real domain thus falling back into the data collection problem.
A more viable solution is to consider a few-shot setting where only a few annotated images from the real target are needed, rather than many target images without annotation.

The few-shot learning problem has been studied in several visual learning scenarios (see \cref{sec:related} for a review of previous works). One of its main challenges is dealing with the intrinsic imbalance between source and target data \cite{sun2019not}. 
When the few-shot learning is considered within the semantic segmentation scenario, this issue is exacerbated by the intrinsic pixel-wise imbalance among segmented classes: 
some classes are both extremely frequent and spatially extended (\eg, sky, road), while others may appear seldom and be small in size (\eg, traffic sign). 
This implies that there can be a great disproportion in the number of pixels-per-class available in the target domain, with some classes that may be scarcely represented or even missing.
This imbalance is more pronounced than in other problem settings, causing image-wise adversarial training methods to align large and well-represented classes, resulting in less accurate mapping of those that are under-represented in the target domain (see \cref{fig:teaser}).

We argue that to address successfully cross-domain few shot learning in semantic segmentation, it is imperative to embed in the solution the intrinsic pixel-wise connotation of spatially segmenting classes. 
To do this, we introduce the Pixel-By-Pixel Cross-Domain Alignment framework ({\our}), that uses a novel pixel-wise discriminator and modulates the adversarial loss for each pixel to: (i) align pixel-wise source and target domains; (ii) avoid to further align correctly represented pixels and reduce the negative transfer; (iii) regularize the training of underrepresented classes to avoid overfitting.
The pixel-wise adversarial training is assisted by a sample selection procedure that handles the imbalance between source and target data by progressively eliminating samples from the source domain. The two mechanisms coexist within an end-to-end training process. Summarizing, the main contributions of this paper are:
\begin{itemize}
    \itemsep -0.2em 
    \item  we propose the first algorithm for cross-domain few shot semantic segmentation able to deal with classes scarcely represented in the training data by spatially aligning the domains pixel by pixel; 
    \item we define a new pixel-wise adversarial loss that aligns source and target domains locally while reducing negative transfer and avoiding overfitting the under-represented classes;
    \item we evaluate our architecture on the two standard synthetic-to-real scenarios, \ie, GTA5$\to$Cityscapes and SYNTHIA$\to$Cityscapes, where it sets new state-of-the-art scores. 
    Additionally, an in-depth ablation study analyzes the influence of all the features introduced by our method. 
\end{itemize}

%------------------------------------------------------------------------
\section{Related Works}
\label{sec:related}
\vspace{-5pt}
\myparagraph{Semantic Segmentation.}
Over the last few years semantic segmentation has achieved remarkable results thanks to the widespread use of deep learning \cite{long2015fully, chen2018encoder, zhao2017pyramid, lin2017refinenet, zhang2018exfuse}. The current state-of-the-art methods differentiate themselves in the strategy applied to condition the semantic information on the global context. Methods like RefineNet~\cite{lin2017refinenet}, PSPNet~\cite{zhao2017pyramid}, ExFuse~\cite{zhang2018exfuse} or DeepLab~\cite{chen2017deeplab, chen2017rethinking, chen2018encoder} are designed to capture objects as well as image context at multiple scales. Other works model the hierarchical or the spatial dependencies to boost the pixel-level classifier \cite{chen2017rethinking,ghiasi2016laplacian}.
One problem with all these methods is that they require a large amount of densely annotated images, which are expensive and time-consuming to obtain. This issue has spurred the creation of synthetic datasets~\cite{tavera2020idda, Ros2016Synthia, Richter2016GTA} that offer high quality images with automatically generated semantic labels. Despite the clear advantages in terms of data availability and quality of the annotations, models trained using synthetic datasets face a drastic domain gap when tested with real images.

\vspace{-5pt}
\myparagraph{Domain Adaptation.}
Domain Adaptation (DA) refers to the study of solutions to bridge the domain gap that is present when the data used to develop the model (source) and the data the model is applied to (target) come from different distributions. Some of these solutions seek to minimize a measure of the discrepancy across domains, like the MMD in \cite{geng2011daml, pmlr-v37-long15}. 
Other methods exploit generative networks and image-to-image translation algorithms to generate target images conditioned on the source domain or vice-versa~\cite{hoffman18cycada,wu2018dcan,yang2020fda}. 
Strategies like~\cite{li2019bidirectional, kim2020learning, yang2020fda} combine image-to-image translation with self-learning, using the predictions on a previously pre-trained model as pseudo labels to fine-tune and reinforce the model itself. 
Finally, the most popular approach for domain adaptation in semantic segmentation is adversarial training~\cite{vu2019advent,luo2019taking,chang2019all}. In the Unsupervised DA setting, Luo \etal \cite{luo2019taking} introduces the negative transfer problem caused by the common global-level adversarial alignment strategy, addressing it with a co-training strategy and an alignment performed at a category-level.
Conversely, we focus on Few-Shot DA for the autonomous driving scenario and propose a novel training strategy that strengthens domain alignment at pixel level while addressing both negative transfer and overfitting on the target domain. 

\vspace{-5pt}
\myparagraph{Few Shot.}
Few-shot learning deals with novel classes given only few images \cite{shaban2017one, rakelly2018few, dong2018few, zhang2019canet, snell2017prototypical, koch2015siamese, finn2017model, hariharan2017low, sung2018learning}. The problem has been extensively studied in the context of image-classification \cite{ snell2017prototypical, koch2015siamese, finn2017model, hariharan2017low,sung2018learning} and, only recently, in the context of semantic segmentation \cite{shaban2017one, rakelly2018few, dong2018few, zhang2019canet}. 
Differently from few-shot learning, the purpose of few-shot in domain adaptation is to transfer the knowledge from a well-annotated source dataset to a target one containing only few annotated images \cite{dong2018few,luo2020adversarial,motiian2017few,zhang2019few}. 
FSDA~\cite{zhang2019few} tackles this problem in semantic segmentation with a two-stage method: the first stage implements a static label filtering that guides the learning towards the pixels that are difficult to classify; the second stage performs domain adaptation at image-level via two image-wise domain discriminators and using all the source images, which forces a negative transfer to the target realm. 
Conversely, our work achieves domain alignment at pixel-level granularity using a new pixel-wise discriminator and a new loss function exploiting the semantic and visual information for each individual pixel. 
Moreover, we use a novel sample selection strategy to limit the number of source images used and avoid a negative transfer.

\myparagraph{Knowledge Distillation.}
Knowledge distillation (KD) \cite{hinton_2015} is applied to transfer knowledge from a cumbersome model to a lighter model with the aim of improving the performance of the latter by forcing the match between the predictions provided by the two networks. It has been applied first to image classification  \cite{NIPS2014_ea8fcd92, Romero2015FitNetsHF, Zagoruyko2017AT} and object detection problems \cite{li_2017}, and only in recent years it has been deployed to the semantic segmentation \cite{Xie2018ImprovingFS, liu2019structured, Shu2020ChannelwiseDF} and the incremental learning tasks \cite{cermelli2020modeling, Douillard2020PLOPLW, michieli2019}. With {\our} we use KD as a regularization term \cite{yuan2020revisiting} to avoid catastrophic forgetting of the acquired knowledge and to avoid overfitting towards the few number of target images provided by the considered few-shot setting.

%-------------------------------------------------------------------------

\begin{figure*}[!t]
\begin{center}
\includegraphics[width=1.0\textwidth]{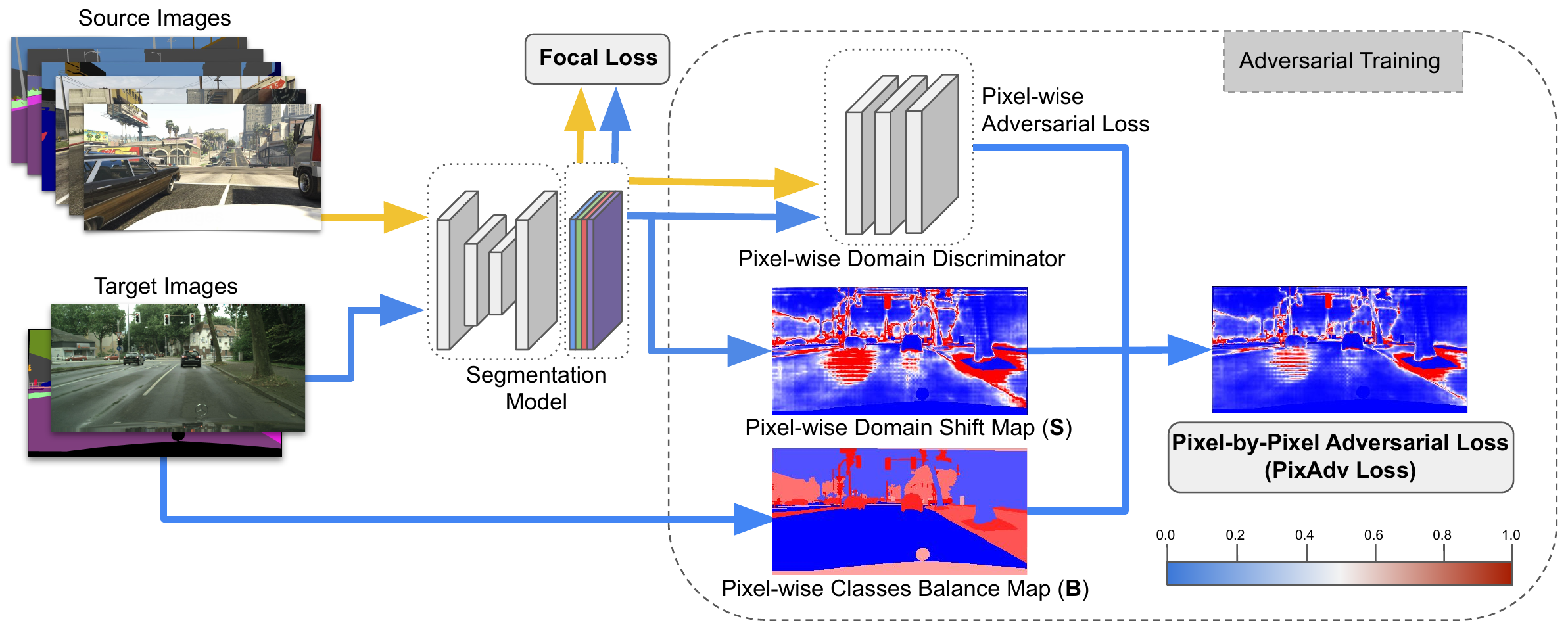}
\end{center}
\caption{Illustration of the pixel-by-pixel adversarial learning of {\our}. A new pixel-wise discriminator computes the adversarial loss whose contribution at each pixel is weighted by two terms: $S$, that considers the ability of the model to correctly represent the pixel, and $B$, that weights each pixel based on the frequency of its semantic class. Yellow/blue lines refer to the source/target domain, respectively.
}
\label{fig:framework}
\vspace{-10pt}
\end{figure*}

\section{Method}
\subsection{Problem Setting} \label{sec:problem}
We consider the semantic segmentation task in the cross-domain few-shot setting that was formulated by Zhang \etal in \cite{zhang2019few}. The problem setting defines $\mathcal{K}$-shot as a task providing $\mathcal{K}$ real images randomly selected for each of the $\mathcal{N}$ cities of the target dataset.
For example, in the 1-shot setting with Cityscapes as target dataset, the whole target data is made of 18 annotated frames because Cityscapes is composed by 18 different cities.
This problem setting is tailored for the autonomous driving application, where a single self-driving solution is usually deployed to a finite number of designated cities.
Although not all the datasets available in the literature provide meta information regarding the division in different cities, this formulation from \cite{zhang2019few} gives a precise and well established experimental protocol. 

To tackle the problem, let us denote as $\mathcal{X}$ the set of RGB images composed by the set of pixels $\mathcal{I}$, and as $\mathcal{Y}$ the set of semantic masks associating to each pixel $i \in \mathcal{I}$ a class from the set of semantic classes $\mathcal{C}$.
At training time we have available two sets of semantically annotated images: $X_{s} = \{(x^{s}, y^{s})\}$  which is a collection of $N_{s}$ images, with $x^{s} \in \mathcal{X}$ from a synthetic domain (\emph{source}), and $X_{t} = \{(x^{t}, y^{t})\}$ which contains a small number of samples $x^{t} \in \mathcal{X}$ from the real-world domain (\emph{target}). Similarly to~\cite{zhang2019few}, the evaluations discussed in Sec. \ref{sec:experiments} are carried out in the (1-5)-shot setting.
%, \ie, $|X_{t}|=\{1-5\}$.
In this notation, $y^{s}, \; y^{t} \in \mathcal{Y}$ denote the annotation masks associated with the source and target images, respectively.
In this problem the goal is to use the datasets $X_s$ and $X_t$ to learn a function $f$, parameterized by $\theta$, from the input space $\mathcal{X}$ to a pixel-wise probability, \ie, $f_\theta: \mathcal{X} \rightarrow \mathbb{R}^{|\mathcal{I}|\times|\mathcal{Y}|}$, and evaluating it on unseen images from the target domain. In the following, we indicate the model output in a pixel $i$ for the class c as $p_i^c$, \ie, $p_i^c(x) = f_\theta(x)[i,c]$. 

Without domain adaptation, the parameters $\theta$ are optimized to minimize the segmentation loss $L_{seg}$:
\begin{equation}
    L_{\text{seg}}(x, y) = - \frac{1}{|\mathcal{I}|} \sum_{i \in \mathcal{I}} (\alpha(1-p^{y_{i}}_i(x))^{\gamma } \log(p^{y_{i}}_i(x))
    \label{eq:focal}
\end{equation}
where $L_{seg}$ corresponds to a focal loss \cite{focal} and $\alpha(1-p^{y_{i}}_i(x))^{\gamma}$ is its modulating factor.

\subsection{Pixel-by-Pixel Adversarial Training} \label{sec:main_method}
Many approaches~\cite{vu2019advent,luo2019taking,chang2019all} in domain adaptation deal with the domain shift problem by aligning the features extracted from the source and target domains in an adversarial manner. The common solution, first introduced by \cite{hoffman2016fcns}, is to play a min-max game between the segmentation network and an image-wise domain discriminator, in which the discriminator predicts the domain a feature belongs to, and the segmentation network tries to deceive it by making source and target features indistinguishable.

Since the domains are analyzed and aligned from a global perspective, the discriminator may disregard portions of the scene that expose few pixels of the small classes, focusing mainly on the well-represented ones. 
As a result, adversarial training would mostly align big and well-represented classes while inducing a negative transfer \cite{luo2019taking} on the others, which leads to poor adaptation.
This problem is amplified in the few shot scenario since there is a discrepancy between the number of images in the source and target domains, and some target semantic classes may be underrepresented or even absent.

\myparagraph{The {\ourloss} Loss.} 
To address the imbalance among classes and reduce the negative transfer, we propose a novel adversarial loss that analyzes each pixel individually rather than operating on a global level (see \cref{fig:framework}).
Our goal is to prioritize and improve pixel alignment using three criteria: (i) align the source and target domain, (ii) avoid to further align correctly represented pixel, limiting negative transfer, and (iii) regularize the training of infrequent classes, forcing the domain alignment to avoid overfitting. 

To accomplish this, we use a pixel-wise discriminator whose goal is to discern, for each pixel, what domain it belongs to. The domain discriminator is a computationally less expensive version of the common Fully Convolutional discriminator found in DCGANs~\cite{radford2016unsupervised} (see  Sec.~\ref{sec:implementation} for more details). 
The discriminator $D$ is trained to classify whether the features are coming from the source or the target domain. Formally, we minimize the following loss:
\begin{equation}
    L_{D}(x^s, x^t) = -\sum_{i\in \mathcal{I}}
     \log\ D_{i}(f_\theta(x^s)) +
     \log(1-D_{i}(f_\theta(x^t))), 
\end{equation}
where $D$ is the discriminator, and $D_{i}(x)$ indicates the output probability for the pixel $i$ to belong to source domain. 

However, using a pixel-wise discriminator without considering the class imbalance problem does not prevent a negative transfer effect. 
Hence, we introduce a novel adversarial loss function (\emph{{\ourloss} Loss}), denoted as $L_{{\ourloss}}$, designed to align each pixel according to its importance.
In particular, to determine the strength with which each pixel is aligned, the $L_{\text{{\ourloss}}}$ modulates the adversarial loss according to a combination of two different terms, each with a specific purpose:
\begin{equation}
\begin{multlined}
    \label{eq:fuse}
    L_{\text{{\ourloss}}}(x^t,y^t) = - \frac{1}{|\mathcal{I}|} \sum_{i \in \mathcal{I}} S_{i}(x^t,y^t) B_{i}(y^t) \log D_{i}(f_\theta(x^t)).
\end{multlined}
\end{equation}

The term $S$ in eq. (\ref{eq:fuse}) is related to the network classification confidence and it is considered as a measure, for each pixel, of the ability of the network to represent it:
\begin{equation} 
    \label{eq:fuse_S}
    S_{i}(x^t,y^t) =  - y_i \;\log p^{y_i}_i(x^t),
\end{equation}
where $p^{y_i}_{i}(x)$ denote the probability for class ${y_{i}}$ at pixel $i$. High values of $S_{i}$ indicate that the network misrepresents the pixel $i$, whereas a small value indicates that the network is able to correctly represent and classify it.

The term $B$ in eq. (\ref{eq:fuse}) represents the imbalance of the pixels and aims to re-balance the classes contribution based on their frequency in the target dataset:
\begin{equation}
    \label{eq:fuse_B}
     B_{i}(y^t) = 1- \frac{1}{{|\mathcal{I}|}} \sum_{j \in \mathcal{I}} \mathbbm{1}_{y_j = y_i},
\end{equation}
where $\mathbbm{1}$ is the indicator function, being one when $y_j$ and $y_i$ are equal, zero otherwise. 
Values of $B_{i}$ which tend to 1 refer to a misrepresented class while values tending to 0 refer to a well-represented class. The term $B$ is crucial since the target domain exposes many pixels of some classes (\eg, road, sidewalk) but very few of others (\eg, train, person).
Through $B$ we are able to balance the classes, resulting in a more heterogeneous and effective adaptation. We would like to point out that the terms $S$ and $B$ aren't used in backpropagation, but rather as a pixel-by-pixel map to modulate the adversarial loss.

Summing up, the overall segmentation network training loss function is expressed as follows: 
\begin{equation}
\begin{multlined}
        \frac{1}{|X_s^k|} \sum_{x^s \in X_s^k} L_{\text{seg}}(x^s, y^s) + \\ \frac{1}{|X_t|} \sum_{x^t \in X_t} L_{\text{seg}}(x^t, y^t) + \lambda L_{\text{{\ourloss}}}(x^t, y^t),
\end{multlined}
\label{scenetotloss}
\end{equation}
where $L_{\text{{\ourloss}}}$ is the proposed adversarial pixel-wise {\ourloss} loss and $X_s^k$ is a subset of the source dataset $X_s$ selected with the sample selection procedure. 

\subsection{Sample Selection} \label{sec:sample_selection}
Due to the extent and variety of the synthetic source dataset there will be source samples far away and detached from the target domain (\eg, different perspective or illumination condition). Forcing the alignment to these samples can result in negative transfer in the target dataset, lowering the network's overall performance. 

With this in mind, we propose a sample selection procedure that, working side by side with the {\ourloss} loss, enhances the use of the source data by identifying and selecting source samples that are better aligned with the target semantic distribution.
Without affecting the segmentation model and not taking part in the adversarial learning process, we simultaneously train a global image-wise domain discriminator $D_{g}$  with the following loss: 
\begin{equation}
\begin{multlined}
    L_{D_g}(x^s, x^t) = - \log\ D_g(f_\theta(x^s)) - \log (1 - D_g(f_\theta(x^t))).
\end{multlined}
\end{equation}

\begin{figure}[t]
    \centering
    \includegraphics[width=1.0\columnwidth]{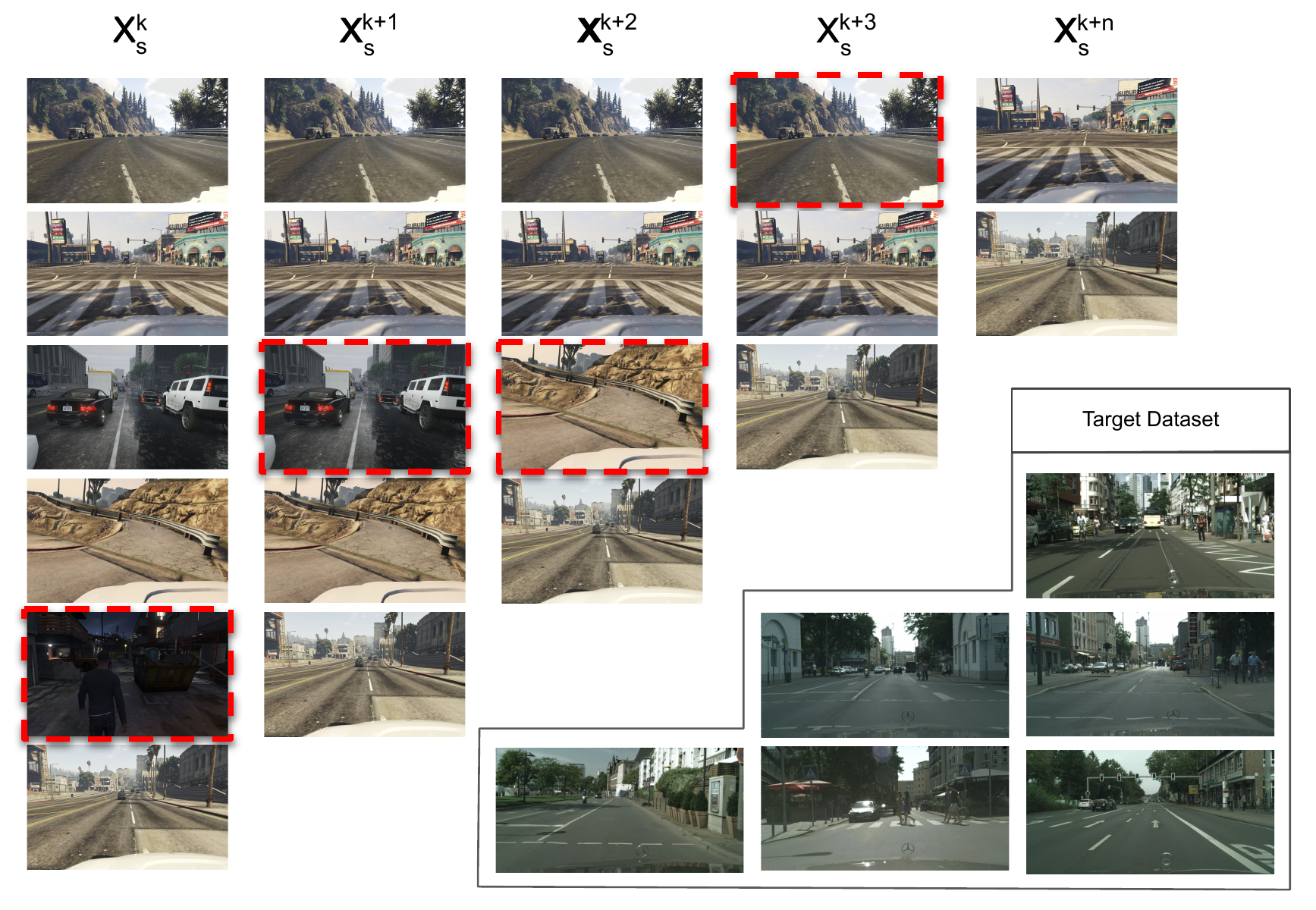}
    \caption{Illustration of the sample selection mechanism (top-left). At each epoch $k$ the source dataset is subsampled selecting images that would carry worthy information to the target domain. For example, samples with different perspectives of lighting conditions w.r.t. the target data (bottom-right) are discarded.}
    \label{fig:sample_selection_example}
    \vspace{-10pt}
\end{figure}

The main reason for using such a discriminator is to distinguish source from target and to capture both semantic and visual domain information. Formally, at each epoch $k$ we exploit $D_{g}$ to predict the likelihood that a source image carries worthy information to the target domain and use this prediction to select a subset $X_s^k$ of source images to be retained from the previous epoch, \ie, $|X_s^{k}| \leq |X_s^{k-1}|$.
Following this intuition, an image $x^s \in X_s^{k-1}$ is added to $X_s^{k}$ if $D_{g}(x^s) < \delta$, where $\delta$ is a predefined threshold. 
After each epoch, we raise the threshold consequently to the increasing capacity of the image-wise discriminator to correctly classify the target data as training progresses, selecting an ever decreasing number of relevant samples (see \cref{fig:sample_selection_example} for a better understanding). 

\subsection{Fine-Tuning and Knowledge Distillation} \label{sec:fine_tuning}
Once the adversarial training is completed and {\ourloss} loss has aligned the representation between pixels of the source and target domains, we can further exploit the available semantic information on the target data to enhance the network representation.

However, na\"ively fine-tuning on the target data ignores the domain alignment obtained previously and may lead to overfitting the few target images. To avoid this problem we use a Knowledge Distillation (KD) strategy. KD \cite{hinton_2015} has been designed to regularize the training of a student network using the output of a teacher network. In our framework the student $f_{\theta_{S}}$ corresponds to the segmentation model which is fine-tuned, 
while the teacher, denoted as $f_{\theta_{T}}$, is a frozen copy of the same network after the adversarial learning process.
Formally, we optimize $\theta_{S}$ with the following: 
\begin{equation}
 \begin{multlined}
    \frac{1}{|X_t|} \sum_{x^t \in X_t} L_{\text{seg}}(x^t, y^t) + \lambda_{kd} L_{\text{kd}}(x^t, f_{\theta_{T}}, f_{\theta_{S}}),
\end{multlined}
\end{equation}
where $L_{seg}$ is the segmentation loss from \cref{eq:focal}, $\lambda_{kd}$ is a weighting parameter. $L_{kd}$ is the distillation loss expressed as follow:
\begin{equation}
 \begin{multlined}
    L_{kd} = - \sigma(\frac{f_{\theta_{T}}(x^t)}{\tau}) \log \sigma(f_{\theta_{S}}(x^t)),
\end{multlined}
\end{equation}
where $\sigma$ indicates the softmax function, 
and $\tau$ is a temperature, as in \cite{hinton_2015}.

%-------------------------------------------------------------------------
\section{Experiments}
\label{sec:experiments}

\subsection{Datasets and metric}
We assess the performance of our method on the two standard synthetic-to-real benchmarks used in the domain adaptation literature: GTA5 \cite{Richter2016GTA} to Cityscapes \cite{Cordts2016Cityscapes}, and SYNTHIA \cite{Ros2016Synthia} to Cityscapes \cite{Cordts2016Cityscapes}. 

\myparagraph{GTA5}. It consists of 24966 images synthesized from the homonymous video game. The original images size is $1914 \times 1052$. For training and evaluation we used the standard 19 semantic classes in common with Cityscapes.

\myparagraph{SYNTHIA}. 
We use the "RAND-CITYSCAPES" subset that consists of 9400 images synthesized from a virtual world simulator. The original image resolution is $1280 \times 760$. The 19 classes in common with Cityscapes are considered for training while the evaluation, following the standard protocol used in \cite{yang2020fda} and in \cite{vu2019advent}, is performed on a subset of 13 and 16 classes.

\myparagraph{Cityscapes}.
It is a real-world dataset collected across several German cities. It consists of 2975 images but for the experiments we use only a subset according to the standard $K$-shot selection ($K$ images from each of the $M$ city). We use the whole validation set made of 500 images to test our network. The original resolution is $2048 \times 1024$. 

We assess the efficiency of {\our} in these two domain adaptation scenarios: GTA5$\to$Cityscapes and SYNTHIA$\to$Cityscapes.
In all tests we use the standard Intersection over Union metric \cite{miou} to measure performance.

\subsection{Implementation and training details} \label{sec:implementation}
\myparagraph{Architecture.}
The segmentation module of our method is DeepLab V2 \cite{chen2017deeplab} with ResNet101 \cite{he2016deep} pre-trained on ImageNet.  The pixel-wise discriminator we built is a Fully Convolutional discriminator which has $2$ convolutional layers with kernel $3 \times 3$, stride 1 and padding 1, followed by a last convolutional layer with kernel $1\times 1$, stride 1 and padding 0. The three layers channel numbers are $\{64, 128, 1\}$. The image-wise discriminator is a common Fully Convolutional discriminator with $5$ convolutional layers with kernel $4\times4$, channel numbers $\{64, 128, 256, 512, 1\}$ and stride 2. For both discriminators, each layer except the last one is followed by a Leaky ReLU with a negative slope of $0.2$. 

\myparagraph{Training.} 
We implement our method in PyTorch and deploy it on two NVIDIA Tesla V100 GPUs with 16GB each. The segmentation model is trained using batch size $4$ and SGD with initial learning rate $2.5 \cdot 10^{-4}$ and adjusted at each iteration with a "poly" learning rate decay with a power of $0.9$, momentum $0.9$ and weight decay to $0.0005$. The discriminators are trained using Adam optimizer, with learning rate $10^{-5}$ and the same decay schedule of the segmentation model. The momentum for Adam is set to \{0.9, 0.99\}.
To reduce the low-level visual domain shift (\eg, color, brightness, etc.) between the source and target domains (both in the adversarial training and in the sample selection phases) we apply to each source image the FFT style translation algorithm from FDA~\cite{yang2020fda}, which is parameterless and computationally light.
{\our} training starts with a pre-trained version of the segmentation model on source data and continues until the sample selection module selects relevant source images for the next epoch. The last fine-tuning and knowledge distillation phase lasts 200 iteration.
We set $\lambda$ equal to $0.1$ for GTA and $1$ for Synthia. The sample selection threshold $\delta$ is set to $0.4$ and doubled at every epoch. Finally, $\lambda_{kd}=0.5$ and $\tau=0.5$. Test is performed without any post-processing.

\myparagraph{Baselines.}
Our method is compared to several baselines. The first baseline that we consider is the Source Only model, \ie, the network trained only with the source dataset. The Joint Training (JT) baseline, that trains for 4 epochs the model with a concatenation of the source and target images. The Fine-Tuning (FT) baseline, that fine-tunes for 30k iterations the Source Only model on the target domain. Our method, JT and FT exploit the Focal Loss to compute segmentation accuracy. 
We then report results for three state-of-the-art methods: FDA \cite{yang2020fda}, NAAE \cite{sun2019not}, and  FSDA \cite{zhang2019few}. FDA \cite{yang2020fda} and "Not All Areas are Equal" (NAAE) \cite{sun2019not} are implemented using the same hyper-parameters proposed in their original papers, replacing only the target train set with the K-shot selection. For FSDA \cite{zhang2019few} we follow the same results and implementation details reported by the authors.
DeepLabV2 with ResNet101 is used as the backbone for all the baselines with the only exception of NAAE that, as provided by its authors, uses a FCN \cite{long2015fully} with VGG16 \cite{vgg}.

\subsection{Results}
\begin{table*}[ht]
\begin{adjustbox}{width=1.0\textwidth}
\centering
\centering
\begin{tabular}{ll|lllllllllll||llllllll|lll}
                        &                & \multicolumn{11}{c||}{\textit{Well}-represented Classes}                                                                                                                                                                                                                                                                                                                                                                 & \multicolumn{8}{c|}{\textit{Under}-represented classes }                                                                                                                                                                                                                                               &         &           &             \\ 
\hline
Shot                   & Method         & \begin{sideways}Road\end{sideways} & \begin{sideways}Building\end{sideways} & \begin{sideways}Vegetation\end{sideways} & \begin{sideways}Sidewalk\end{sideways} & \begin{sideways}Car\end{sideways} & \begin{sideways}Sky\end{sideways} & \begin{sideways}Pole\end{sideways} & \begin{sideways}Terrain\end{sideways} & \begin{sideways}Person\end{sideways} & \begin{sideways}Fence\end{sideways} & \begin{sideways}Wall\end{sideways} & \begin{sideways}TSign\end{sideways} & \begin{sideways}Bicycle\end{sideways} & \begin{sideways}Bus\end{sideways} & \begin{sideways}Truck\end{sideways} & \begin{sideways}Train\end{sideways} & \begin{sideways}TLight\end{sideways} & \begin{sideways}Rider\end{sideways} & \begin{sideways}Motorcycle\end{sideways} & mIoU$^{19}$ & mIoU$^{Well}$ & mIoU$^{Under}$  \\
\end{tabular}
\end{adjustbox}
\vspace{1mm}
\caption{GTA5$\to$Cityscapes Experiments. Classes are sorted by decreasing frequency on target domain.}
\label{table:gta}
\vspace{7pt}
\end{table*}

\vspace{-5pt}
\myparagraph{GTA5 to Cityscapes.}
The results for this scenario are reported in Table \ref{table:gta}. 
At a first glance, we observe that NAAE and Joint Training lead to underwhelming results, with a mIoU below $40\%$ in all tests. FDA is slightly better, but its accuracy does not improve when increasing the number of target images from 1 to 5.
Fine-Tuning the model pretrained on the source domain leads to comparable accuracy to the current state-of-the-art, FSDA. 
Finally, our {\our} is the best performer in all (1-5)-shot tests, outperforming the Source Only model by a minimum of $+20.44\%$ in the 1-shot setting to a maximum of $+24.84\%$ in the 5-shot.

Compared to the next best competitor, \ie, FSDA, {\our} marks an average boost of $+3.63\%$ to the mIoU.
We also note that in the 1-shot setting, the accuracy of FSDA in few classes (traffic light, motorcycle) drops below the Source Only baseline, which is indicative of a negative transfer. This result confirms that {\our} uses more effectively the information from the domain images depending on the content in the target images.

\begin{figure*}[t]
\centering
\includegraphics[width=1.0\textwidth]{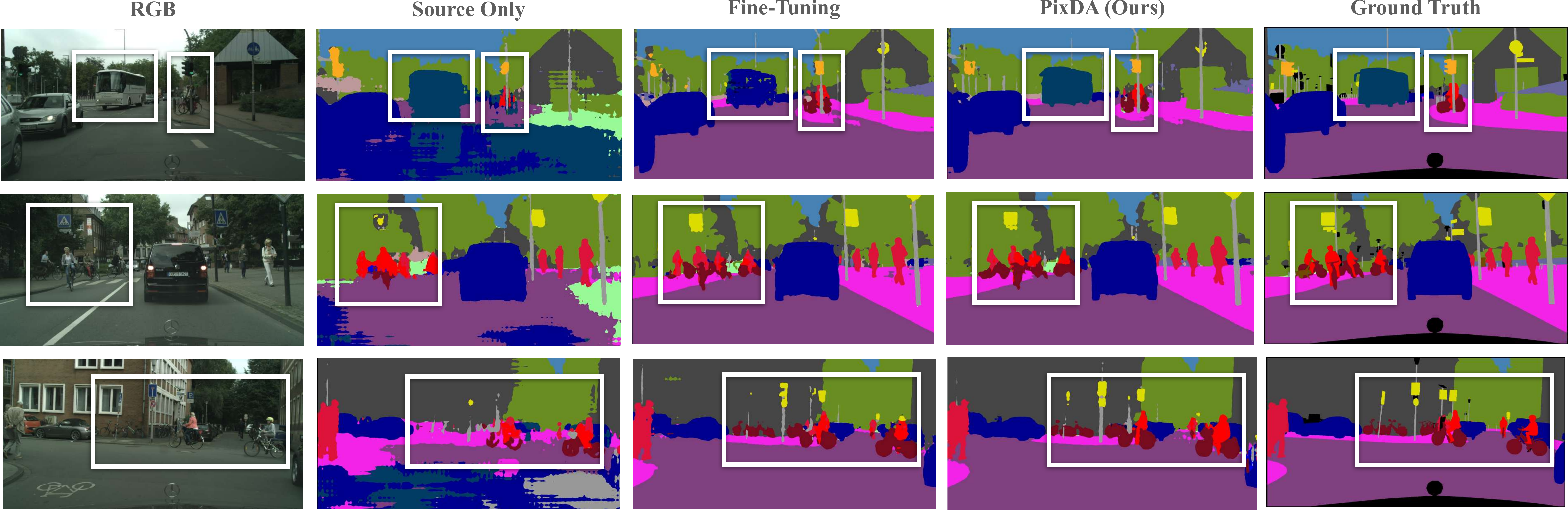}
\caption{Qualitative results for the GTA$\to$Cityscapes and SYNTHIA$\to$Cityscapes scenarios.
The boxes on the images highlight some examples of underrepresented classes. Qualitative comparison to the other methods is left to the Supplementary Material.}
\label{fig:qualitatives}
\end{figure*}

Finally, we observe that our method not only works well with predominant classes such as "road", "sky" and "building", but on average it improves the recognition of semantic categories that are under-represented, either because containing few pixels (\eg, "traffic light", where we achieve a $+9.39\%$ \wrt to the Source Only) or because rarely appearing (\eg, "train", where we achieve $+15.51\%$ \wrt to the Source Only). 
Overall, on under-represented classes (last column in \cref{table:gta}) we outperform FSDA by $+6.58\%$, demonstrating our ability to correctly align the pixels related to these categories.
These results are qualitatively confirmed in Fig. \ref{fig:qualitatives}, where we show that the {\ourloss} loss provides a stronger adaptation for small and rare classes, such as "traffic sign" and "bicycle"; hence, these categories are predicted quite accurately even in the 1-shot setting.

\begin{table*}[ht]
\begin{adjustbox}{width=1.0\textwidth}
\centering

\begin{tabular}{ll|llllllllll||llllll|llll}
                        &                & \multicolumn{10}{c||}{\textit{Well}-represented Classes}                                                                                                                                                                                                                                                                                                                         & \multicolumn{6}{c|}{\textit{Under}-represented Classes}                                                                                                                                                                   &              &              &           &             \\ 
\hline
Shot                    & Method         & \begin{sideways}Road\end{sideways} & \begin{sideways}Building\end{sideways} & \begin{sideways}Vegetation\end{sideways} & \begin{sideways}Sidewalk\end{sideways} & \begin{sideways}Car\end{sideways} & \begin{sideways}Sky\end{sideways} & \begin{sideways}Pole\end{sideways} & \begin{sideways}Person\end{sideways} & \begin{sideways}Fence\end{sideways} & \begin{sideways}Wall\end{sideways} & \begin{sideways}TSign\end{sideways} & \begin{sideways}Bicycle\end{sideways} & \begin{sideways}Bus\end{sideways} & \begin{sideways}TLight\end{sideways} & \begin{sideways}Rider\end{sideways} & \begin{sideways}Motorcycle\end{sideways} & mIoU$^{13}$  & mIoU$^{16}$  & mIoU$^{Well}$ & mIoU$^{Under}$  \\
\end{tabular}

\end{adjustbox}
\vspace{1mm}
\caption{SYNTHIA$\to$Cityscapes Experiments. Classes are sorted by decreasing frequency on target domain.}
\label{table:synthia}
\vspace{-5pt}
\end{table*}

\myparagraph{SYNTHIA to Cityscapes.}
Results for this scenario are reported in Table \ref{table:synthia} and shown in \cref{fig:qualitatives} and confirm what we showed in the first set of experiments.
NAAE, FDA and Joint Training lag behind all other methods, confirming that they are not viable solutions to address the cross-domain few-shot problem.
Fine-Tuning and FSDA show similar accuracies, although it must be noted that the results from FSDA are only reported for the protocol with 13 classes, where the difficult categories "pole", "fence" and "wall" are excluded from the evaluation. Our method is designed to better handle such categories as well and, on these classes, it improves its performance \wrt the Source Only model by $+22.88\%$, $+13.60\%$ and $+15.16\%$ respectively.
Overall, {\our} displays the best mIoU in all (1-5)-shot settings, both with the 13 and 16 classes protocols. It outperforms the Source Only model by a minimum of $+25.72\%$ in the 1-shot scenario to a maximum of $31.66\%$ in the 5-shot. 
Compared to the current state-of-the-art, \ie, FSDA, {\our} scores an average accuracy improvement of $+1.69\%$ within the 13 classes protocol and of $+1.86\%$ if considering only the rare classes. 
The closest method to {\our}, \ie, Fine-Tuning, achieves good results on these three categories but it is much less consistent than our solution.
In the 16-classes protocol, {\our} achieves an average boost of $+2.56\%$ \wrt the second best competitor, \ie, Fine-Tuning.

%-------------------------------------------------------------------------
\subsection{Ablation Study} \label{sec:ablation}
\vspace{-2mm}
\myparagraph{Contribution of terms in the PixAdv loss.}\label{sec:fuse_loss_ablation}
In \cref{table:loss_ablation} we provide an in-depth ablation study to prove the effectiveness of making domain alignment at a pixel-level. The results are computed while both sample selection and knowledge distillation and fine-tuning are turned off.
The table shows that aligning source and target domains with an image-wise discriminator yields a substantial improvement ($+11.14\%$) over joint-training (no adversarial loss). However, since it discriminates the images globally, it tends to align well-represented classes while ignoring the others.
In comparison, the pixel-level adversarial loss, which aligns each pixel separately, further enhances performance by $+2.17\%$. Regardless of the change, merely aligning the pixels does not prevent negative transfer and overfitting to few-shot images. Indeed, re-weighting the pixels based on their frequency ($B$ term) yields a further improvement of $+1.46\%$.
Furthermore, decreasing the weight of well-represented pixels ($S$ term) to prevent negative transfer is crucial, providing a $+1.64\%$ w.r.t the pixel-wise adversarial loss. 

Finally, the $B$ and $S$ terms complement each other and boost efficiency even further when used together.
The resulting loss ({\ourloss}) outperforms the image-wise adversarial loss by $+5.86\%$ and the pixel-wise adversarial loss by $+3.69\%$, suggesting that weighting each pixel contribution is advantageous to prevent negative transfer and overfitting.

\begin{table}[t]
\centering
\setlength{\tabcolsep}{12pt} 

\resizebox{0.85\columnwidth}{!}{
\begin{tabular}{c|c|c|c|l}
Image-wise & Pixel-wise &  &  &  \\ 
Adv. Loss & Adv. Loss & B & S & mIoU \\
\hline
 & & & & 31.29 \\ 
\cmark & & & & 42.43 \\ 
& \cmark & & & 44.60 \\
& \cmark & \cmark &   & 46.06 \\
& \cmark & & \cmark & 47.70 \\
& \cmark & \cmark &  \cmark & \textbf{48.29} \\
\end{tabular}
}
\vspace{1.1mm}

\caption{Ablation study about the choice of the adversarial loss on the GTA$\to$Cityscapes 1-shot scenario.}
\label{table:loss_ablation}
\vspace{-17pt}
\end{table}

\myparagraph{Contribution of each component in PixDA.}
\vspace{-1pt}
In this section we assess to what extent each component in our framework contributes to the final performance. We move bottom-up examining six different cases: (a) Source Only model; (b) Joint Training; (c) training with the {\ourloss} loss; (d) with the {\ourloss} Loss and our Sample Selection mechanism; (e) the fine-tuning step and finally, (f) the knowledge distillation that completes the {\our} framework. 

From the results in \cref{table:da}, it is evident that each component brings an improvement to the overall framework. In particular, the addition of the {\ourloss} Loss improves the Joint Training by $+17\%$, indicating that domain alignment is necessary to obtain good performance.
The Sample Selection provides an additional improvement of $+1.45\%$, indicating that removing samples far from the target distribution is beneficial. Finally, while na\"ively fine-tuning the network on the target brings a little improvement ($+0.31\%$), using knowledge distillation brings that difference to $+1.42\%$. We remark that only adding the {\ourloss} Loss already surpasses the state-of-the-art. As a follow-up test, we replaced the Focal Loss with a standard Cross Entropy, yielding a lower but still state-of-the-art result ($48.89\%$) and confirming the effectiveness of our loss. Additional studies to asses the impact of hyperparameters in our %{\our} 
framework are included in the supplementary material due to lack of space.
\begin{table}[t]
\centering
\setlength{\tabcolsep}{12pt}
\resizebox{.58\columnwidth}{!}{
\begin{tabular}{l|l}
Method & mIoU \\ 
\hline
Source Only                 & 30.72 \\
Joint Training              & 31.29 \\
{\ourloss}                  & 48.29 \\
+ Sample Selection          & 49.74 \\
+ Fine-Tuning               & 50.05 \\
+ KD                        & \textbf{51.16}
\end{tabular}
}
\vspace{1.1mm}
\caption{Ablation study showing the effectiveness of each {\our} component on the GTA$\to$Cityscapes 1-shot scenario.}
\vspace{-12pt}
\label{table:da}
\end{table}

%-------------------------------------------------------------------------
%\input{Sections/6-UDA}

%-------------------------------------------------------------------------
\vspace{-5pt}
\section{Conclusion}
\vspace{-5pt}
In this work we address the task of Cross-Domain Few-Shot Semantic Segmentation. We present a pixel-by-pixel adversarial training strategy that uses a new pixel-wise loss and discriminator to better align source and target domain and to reduce the negative transfer problem. We also assist the adversarial training with a sample selection procedure that handles the imbalance between source and target domain. Our framework achieves the state-of-the-art performance in all the 1-to-5 shot settings from the two standard synthetic-to-real benchmarks.

Future work will analyze a modified version of the sample selection strategy that selects only the top $K$ confident source samples rather than increasing the threshold, as well as the application of the {\ourloss} Loss to other settings, such as Unsupervised DA, which has a preliminary assessment in the Supplementary Material, and Multi Source DA.

{\small
\bibliographystyle{ieee_fullname}
\bibliography{egbib}
}

\end{document}